\documentclass{article}

     \usepackage[preprint]{neurips_2020}

\usepackage[utf8]{inputenc} 
\usepackage[T1]{fontenc}    
\usepackage{hyperref}       
\usepackage{url}            
\usepackage{booktabs}       
\usepackage{amsfonts}       
\usepackage{nicefrac}       
\usepackage{microtype}      
\usepackage{amsmath}
\usepackage{enumerate}

\newcommand{\x}{\mathbf{x}}
\newcommand{\bb}{\mathbf{b}}

\newcommand{\ba}{\mathbf{a}}
\newcommand{\z}{\mathbf{z}}
\DeclareMathOperator*{\argmax}{arg\,max}

\title{Using noise resilience for ranking generalization of deep neural networks}

\author{%
  Depen Morwani\\
  Department of Computer Science\\
  Indian Institute of Technology Madras\\
  Chennai, 600036 \\
  \texttt{cs19s002@smail.iitm.ac.in} \\
  \And 
    Rahul Vashisht\\
  Department of Computer Science\\
  Indian Institute of Technology Madras\\
  Chennai, 600036 \\
  \texttt{cs18d006@smail.iitm.ac.in} \\
  \And
    Harish G. Ramaswamy\\
  Department of Computer Science\\
  Indian Institute of Technology Madras\\
  Chennai, 600036 \\
  \texttt{hariguru@cse.iitm.ac.in} \\
}

\begin{document}

\maketitle

\begin{abstract}
Recent papers have shown that sufficiently overparameterized neural networks can perfectly fit even random labels. Thus, it is crucial to understand the underlying reason behind the generalization performance of a network on real-world data. In this work, we propose several measures to predict the generalization error of a network given the training data and its parameters. Using one of these measures, based on noise resilience of the network, we secured 5th position in the predicting generalization in deep learning (PGDL) competition at NeurIPS 2020.
\end{abstract}

\section{Introduction}
Deep learning models have achieved tremendous success in a wide variety of tasks related to computer vision (CV) and natural language processing (NLP). However, we still do not have a clear understanding regarding the generalization behavior of these models. Generalization is defined as the ability of classifiers to perform well on unseen data. It is desirable for real world deployment of deep models in real-world applications such as healthcare. Recently, \citet{GoodFellow15} showed that neural networks can misclassify an example even with perturbations imperceptible to human eye, and \citet{Zhang17} demonstrated that sufficiently overparameterized neural networks can even fit random labels. Both of these observations question the ability of a neural network to generalize well. Moreover, in both the cases, cross entropy loss on the training dataset remained small, indicating that it cannot be a predictor of generalization. Thus it is crucial to understand the reason behind generalization of neural networks.

A lot of generalization bounds have been proposed for deep networks based on margin and weight norms \citep{Bartlett17,Pitas17}. In a recent work, \citet{Jiang20} studied the capability of these bounds to predict the generalization gap of a network. In this work, we propose a few other measures based on noise stability and hidden layer margins of the network. 


\section{Notation}
A neural network is represented as a function $f$ that is given by $f(\x) = W_l(\sigma(W_{l-1}...(W_1\x + \bb_1) + \bb_2)....) + \bb_l$, where $l$ represents the number of layers in the network, $\x \in \mathbb{R}^d$ represents the input to the network, $\sigma$ represents the non-linearity in the network operating element-wise and $W_j \in \mathbb{R}^{h_j \times h_{j-1}}, \bb_j \in \mathbb{R}^{h_j}$ represent the weights and biases of the network with $h_j$ representing the width of the $j^{th}$ hidden layer. This representation suffices for a convolutional network as well, because convolution operation can be replaced by a sparse weight matrix. The pre-activations and activations at the $j^{th}$ layer are represented by $\z_j(\x)$ and $\ba_j(\x)$ respectively. Using this notation, the neural network can be defined as
\begin{align*}
    \ba_0(\x) &= \x \\
    \z_{j}(\x) &= W_{j} \ba_{j-1}(\x) + \bb_j \\
    \ba_j(\x) &= \sigma(\z_j(\x))
\end{align*}
$\| W \|_2$ and $\| W \|_\text{F}$ represent the spectral and frobenius norm respectively, of a matrix $W$.
The training dataset is represented by $\mathcal{D} = \{(\x_i, y_i)\}_{i=1}^n$, where $n$ represents the number of points in the training dataset. Let the set of all classes in the dataset be denoted by $\mathcal{Y}$.
$\mathcal{N}(\mathbf{\mu}, \Sigma)$ represents the normal distribution with mean $\mathbf{\mu}$ and covariance $\Sigma$. $\mathbb{E}(\mathbf{X})$ represents the expectation of a random variable $\mathbf{X}$.

\section{Methodology}
The measures used can be broadly divided into three different categories: Noise-based, Margin-based and Norm-based\footnote{Code available: \href{https://github.com/VASHISHT-RAHUL/Generalization_performance_of_neural_networks}{https://github.com/VASHISHT-RAHUL/Generalization\_performance\_of\_neural\_networks}}. In this section, we will provide the description of these methods and their results in the PGDL competition will be discussed in the following section.

\subsection{Noise-based measures}
The noise stability of the $j^{th}$ layer with respect to $(j-1)^{th}$ layer is measured by adding Gaussian noise to the activations of $(j-1)^{th}$ layer and quantifying its propagation to the pre-activations of the $j^{th}$ layer. Precisely, let $\mathbf{Y} \sim \mathcal{N}(\mathbf{0}, I)$, where $\mathbf{Y} \in \mathbb{R}^{h_{j-1}}$. Then, define
\begin{align*}
    \ba'_{j-1}(\x) &= \ba_{j-1}(\x) + \sqrt{\frac{\nu}{d_{j-1}}}\| \ba_{j-1}(\x) \| \mathbf{Y} \\
    \z'_{j}(\x) &= W_{j} \ba'_{j-1}(\x) + \bb_j
\end{align*} 
where $\nu$ controls the standard deviation of the Gaussian noise and the other factors ensure that
$\mathbb{E}(\| \ba'_{j-1}(\x) \|^2) = (1 + \nu)\mathbb{E}(\| \ba_{j-1}(\x) \|^2)$, thus making the noise proportional to the scale of the inputs. In this case, noise stability ($\beta_j(\x)$) of layer $j$ with respect to layer $j-1$ is defined as
\[ \beta_j(\x) = \frac{1}{\nu} \frac{\| \z'_j(\x) - \z_j(\x) \|^2}{\| \z_j(\x) \|^2} \]
Two measures defined using $\beta_j(\x)$ are given below
\[ \textbf{mean-noise-stability} = \frac{1}{n*l}\sum_{i=1}^n \sum_{j=1}^l  \beta_j(\x_i) \]
\[ \textbf{geometric-mean-noise-stability} = \frac{1}{n*l}\sum_{i=1}^n \sum_{j=1}^l  \log(\beta_j(\x_i)) \]
Two other measures can be defined similarly (\textbf{mean-noise-stability-output} and \textbf{geometric-mean-noise-stability-output}), using the noise stability of the output with respect to the input. Variations of this measure have been used in recent papers \citep{Arora18,Nagarajan19} for establishing data-dependent generalization bounds.

\subsection{Margin-based measures}
Multiple generalization bounds based on the output-layer margin have been established for deep networks \citep{Jiang20}. However, margin can be defined at intermediate layers as well \citep{Elsayed18}. Moreover, a linear parametric model with these margins as inputs has been shown to predict the generalization gap reasonably well, given a constant depth \citep{Jiang19}. However, with varying depth, determining the inputs for the parametric model is tricky and moreover, we explicitly wanted to avoid parametric models and focus on theoretically grounded measures. 

\subsubsection{Input-layer margin ($\gamma_{\text{inp}}$)} 
    
It is defined as the minimum perturbation that needs to be added to an input $\x$ so that the perturbed input is misclassified by the network \citep{Elsayed18}. It was approximated using the first-order Taylor expansion of the network around the input.
    
\subsubsection{All-layer-margin ($\gamma_{\text{all}}$)}
    
Consider a neural network represented as a composition of $l$ functions denoted by $f_1, ..., f_l$. Let $\delta_1, ..., \delta_l$ represent the perturbations applied at each layer. Then the perturbed network output ($F(\x, \delta)$) is given by
\begin{align*}
    g_1(\x, \delta) &= f(\x) + \delta_1 \| \x \|_2 \\
    g_j(\x, \delta) &= f_j(g_{j-1}(\x, \delta)) + \delta_j \| g_{j-1}(\x, \delta) \| \\
    F(\x, \delta) &= g_l(\x, \delta)
\end{align*}
Then, all-layer-margin \citep{Wei20} is defined as
\begin{align*}
    \gamma_{\text{all}}(\x, y) =& \min_{\delta_1, ..., \delta_l} \sqrt{\sum_{j=1}^l \| \delta_j \|^2}\\
    &\text{subject to} \argmax_{y' \in \mathcal{Y}} F(\x, \delta)_{y'} \neq y
\end{align*}
It was approximated by performing gradient ascent on the loss function with respect to the parameters $\delta_1, ..., \delta_l$.
    
\subsubsection{Margin-jacobian}
    
This measure was defined by combining the output-layer margin ($\gamma_{\text{out}}$) with the Jacobian of the output with respect to the intermediate layers.
\[ \textbf{margin-jacobian} = \left(\frac{l}{\gamma_{\text{out}}^2}\right)^{\frac{1}{l}} + \frac{\sum_{i=1}^{n} \sum_{j=1}^l \frac{1}{d_l*d_j}\| \frac{\partial f(\x_i)}{\partial \ba_j} \|_{\text{F}}^2}{nl^2\gamma_{\text{out}}} \]
The first term normalizes the margin by the depth and was inspired from one of the bounds in \citet{Jiang20}. $\frac{\sum_{i=1}^{n} \sum_{j=1}^l \frac{1}{d_l*d_j}\| \frac{\partial f(\x_i)}{\partial \ba_j} \|_{\text{F}}^2}{nl}$ represents the average Jacobian norm, that is further normalized by depth and margin. We found that the sum of these terms works better empirically than the individual terms.

\subsection{Norm-based Measures}
Multiple bounds based on spectral and frobenius norms have been proposed recently for feed-forward as well as convolutional networks \citep{Bartlett17,Pitas17}. The capability of these bounds to predict the generalization gap has also been explored in \citet{Jiang20}. We provide an efficient way of evaluating the spectral norm of weight matrices in case of convolutional networks.

Converting the weight matrix of a convolution operator to the sparse weight matrix of a linear operator is time-consuming task. Instead, for calculating the spectral norm, or any eigenvalues of the convolution operator, we used the Power Method, that works for any linear operator, whether or not it is defined explicitly by a weight matrix (for more details, refer to the code). This trick allowed us to evaluate these bounds for deep networks within the time limit of the competition.

Based on this faster method of evaluation, the one metric based on spectral norm that we tried was
\[ \textbf{fast-log-spec} = (1 - \frac{1}{l})\sum_{i=1}^l \log(\| W_i \|_2^2) - \log(\gamma_{\text{out}}^2) \]
This is based on the generalization bound for convolutional networks from \citet{Pitas17}. 

\section{Results in PGDL}
The PGDL competition had three different kind of datasets - public, development and private. Public dataset had two different tasks - Task 1 and Task2, and was completely accessible to the competitors, i.e, we could access the final test error of the networks as well as the final scores on these tasks. Development dataset had two different tasks - Task 4 and Task 5, and was partially accessible, i.e, we could just access the final scores on these tasks. Private dataset had 4 tasks - Task 6, Task 7, Task 8 and Task 9, and was significantly more complex as compared to public and development datasets. Participants were allowed limited submissions on this dataset and the final score was only revealed after the competition ended.

Each task consisted of a dataset and a set of neural networks with varying hyperparameters such as depth, dropout etc., that were trained to almost 100\% accuracy on the training dataset. The aim was to come up with a measure that can predict the test error of the network, given the training data and its parameters. The evaluation metric used was the mutual information between the predicted measure and the generalization error, conditioned on the hyperparameters. More details regarding the competition can be found on the \href{https://sites.google.com/view/pgdl2020}{PGDL} website. 

The results for development and public dataset for all the measures (except for one that we did not try on the development dataset) are given in Table \ref{tab:pubdev}. As limited submissions were allowed on private dataset, results for a few of these measures on the private dataset are shown in Table \ref{tab:priv} (for exact hyperparameter settings, refer to the code). 

As can be clearly seen, \textbf{geometric-mean-noise-stability} clearly beats the other measures on the private dataset. However, its performance on a particular task - Private Task 8 is poor. Similarly, on development and public datasets, it doesn't perform as well as other measures. This clearly indicates a non-uniformity in the measure that is predictive of generalization on a particular task.

\begin{table}[t]
\begin{center}
    \begin{tabular}{ p{5cm} p{1cm} p{1cm} p{1cm} p{1cm} p{1cm} p{1cm} }
    \hline
    Method & Public & Dev & Public Task 1 & Public Task 2 & Dev Task 4 & Dev Task 5 \\
    \hline
    mean-noise-stability & 10.77 & 0.58 & 0.37 & 21.17 & 0.20 & 0.95 \\
    \hline
    geometric-mean-noise-stability & 2.66 & 1.45 & 0.10 & 5.23 & 1.50 & 1.40 \\
    \hline
    mean-noise-stability-output & 6.21 & 1.64 & \textbf{11.07} & 1.35 & 0.27 & 3.02 \\
    \hline
    geometric-mean-noise-stability-output & 5.99 & & 9.53 & 2.45 & &\\
    \hline
    input-layer-margin & 1.93 & 7.36 & 2.63 & 1.22 & 5.36 & \textbf{9.37} \\
    \hline
    all-layer-margin & \textbf{12.79} & 3.30 & 1.23 & \textbf{24.36} & 0.32 & 6.27\\
    \hline
    margin-jacobian & 2.81 & \textbf{9.38} & 0.15 & 5.47 & \textbf{18.23} & 0.53\\
    \hline
    fast-log-spec & 5.95 & 2.11 & 7.04 & 4.87 & 0.05 & 4.17\\
    \hline
    \end{tabular}
    \caption{Results on public and development dataset}
    \label{tab:pubdev}
\end{center}
\end{table}

\begin{table}[t]
\begin{center}
    \begin{tabular}{ p{5cm} p{1cm} p{1cm} p{1cm} p{1cm} p{1cm} }
    \hline
    Method & Private & Private Task 6 & Private Task 7 & Private Task 8 & Private Task 9\\
    \hline
    geometric-mean-noise-stability & \textbf{6.51} & 7.76 & \textbf{8.02} & 0.67 & \textbf{9.58}\\
    \hline
    mean-noise-stability-output & 2.10 & 0.07 & 0.80 & \textbf{5.77} & 1.75\\
    \hline
    input-layer-margin & 1.34 & 1.18 & 2.74 & 0.62 & 0.82\\
    \hline
    all-layer-margin & 2.50 & 3.15 & 2.21 & 0.63 & 4.01\\
    \hline
    margin-jacobian & 4.01 & \textbf{8.50} & 2.02 & 2.67 & 2.84\\
    \hline
    \end{tabular}
    \caption{Results on private dataset}
    \label{tab:priv}
\end{center}
\end{table}

\section{Conclusion}
Although noise resilience based measure beats other measures on the private dataset, the same does not hold for public and development datasets. This clearly indicates a non-uniformity in the measure required to predict generalization for a given task, that can be explored further, if the data of these tasks is made available to the competitors.
\bibliographystyle{unsrtnat}
\small
\bibliography{References}





\end{document}